\documentclass[10pt,twocolumn,letterpaper]{article}

\usepackage{iccv}
\usepackage{times}
\usepackage{epsfig}
\usepackage{graphicx}
\usepackage{amsmath,bm}
\usepackage{amssymb}

\usepackage{booktabs}
\usepackage{xcolor,colortbl}
\definecolor{Gray}{gray}{0.9}
\usepackage{multirow}
\usepackage{multicol}
\usepackage{caption}
\usepackage{subcaption}
\usepackage{times}
\usepackage{cases}
\usepackage{diagbox}
\usepackage{algorithm}
\usepackage{algorithmic}
\usepackage{tablefootnote,footnotehyper}
\newcommand\blfootnote[1]{%
  \begingroup
  \renewcommand\thefootnote{}\footnote{#1}%
  \addtocounter{footnote}{-1}%
  \endgroup
}

\usepackage[accsupp]{axessibility}  

\definecolor{softgreen}{RGB}{102,204,102} 
\usepackage[colorlinks,
            anchorcolor=softgreen,
            citecolor=softgreen,
            bookmarks=false
            ]{hyperref}

\iccvfinalcopy 

\def\iccvPaperID{****} 

\ificcvfinal\fi

\begin{document}

\title{UCF: Uncovering Common Features for Generalizable Deepfake Detection}

\author{
Zhiyuan Yan$^{*1}$
\qquad Yong Zhang$^{*2}$
\qquad Yanbo Fan$^{2}$ 
\qquad Baoyuan Wu$^{\dagger1}$\\
$^{1}$
The School of Data Science, \\
The Chinese University of Hong Kong, Shenzhen (CUHK-Shenzhen), China \\
$^{2}$Tencent AI Lab \\
{\tt\small \{yanzhiyuan1114, zhangyong201303, fanyanbo0124\}@gmail.com, wubaoyuan@cuhk.edu.cn
}
}


\maketitle
\ificcvfinal\thispagestyle{empty}\fi

\begin{abstract}
Deepfake detection remains a challenging task due to the difficulty of generalizing to new types of forgeries. This problem primarily stems from the overfitting of existing detection methods to forgery-irrelevant features and method-specific patterns.
The latter has been rarely studied and not well addressed by previous works.
%
This paper presents a novel approach to address the two types of overfitting issues by uncovering common forgery features. 
Specifically, we first propose a disentanglement framework that decomposes image information into three distinct components: forgery-irrelevant, method-specific forgery, and common forgery features. 
To ensure the decoupling of method-specific and common forgery features, a multi-task learning strategy is employed, including a multi-class classification that predicts the category of the forgery method and a binary classification that distinguishes the real from the fake. 
Additionally, a conditional decoder is designed to utilize forgery features as a condition along with forgery-irrelevant features to generate reconstructed images.
%
Furthermore, a contrastive regularization technique is proposed to encourage the disentanglement of the common and specific forgery features. 
Ultimately, we only utilize the common forgery features for the purpose of generalizable deepfake detection.
Extensive evaluations demonstrate that our framework can perform superior generalization than current state-of-the-art methods. 

\end{abstract}

\blfootnote{$^*$Equal contribution}
\blfootnote{$^\dagger$Corresponding Author}

\begin{figure}[htb]
      \centering 
      \includegraphics[width=1.0\linewidth]{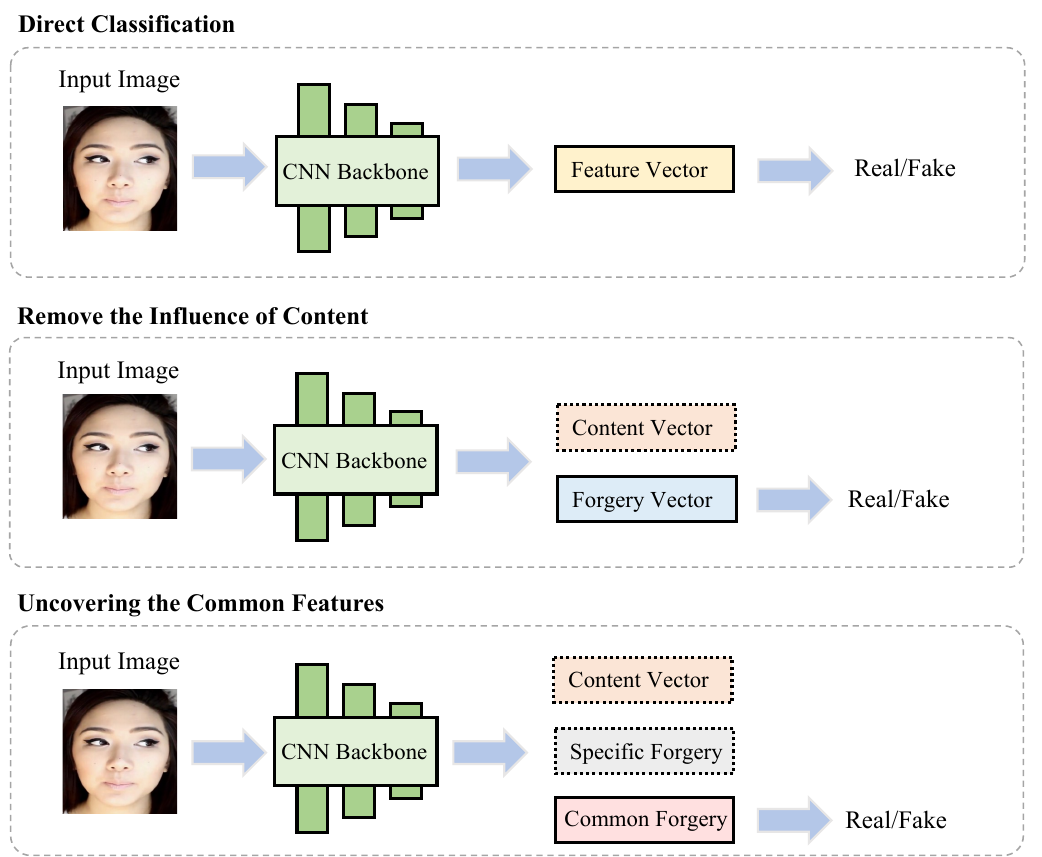} 
      \caption
      {
       Comparison among different classification methods. The first is a direct classification that uses whole features. The second approach eliminates content features to prevent overfitting to forgery-irrelevant features. Our approach, the third one, not only removes the influence of content but also prevents overfitting to specific forgery patterns by uncovering common features.
      }
      \label{concept}
\vspace{-10pt} 
\end{figure}

\vspace{-20pt} 
\section{Introduction}

Deepfake technology has gained significant attention in recent years due to its ability to generate highly realistic videos. While deepfake has the potential to be used for various purposes, including entertainment and marketing, it has also been misused for illegal purposes. The use of deepfake to create false content can compromise people's privacy, spread misinformation, and erode trust in digital media, resulting in severe outcomes like reputational harm, incitement of violence, and political instability.

As a result, developing a reliable and effective deepfake detection algorithm is vitally essential. Recently, a large number of detectors~\cite{zhou2017two,li2018exposing,rossler2019faceforensics++,yang219exposing,qian2020thinking,zhao2021multi,chen2022ost} have been proposed for deepfake. Existing detectors generally perform well when the training and testing data are created using the same forgery techniques. However, in real-world applications, the testing data may be created using unknown procedures, leading to differences between the training and testing data and resulting in poor detection performance. This phenomenon, known as the generalization problem in deepfake detection, presents a significant challenge to the practical use of current detection methods.

\begin{figure*}[htb]
      \centering 
      \includegraphics[width=0.90\linewidth]{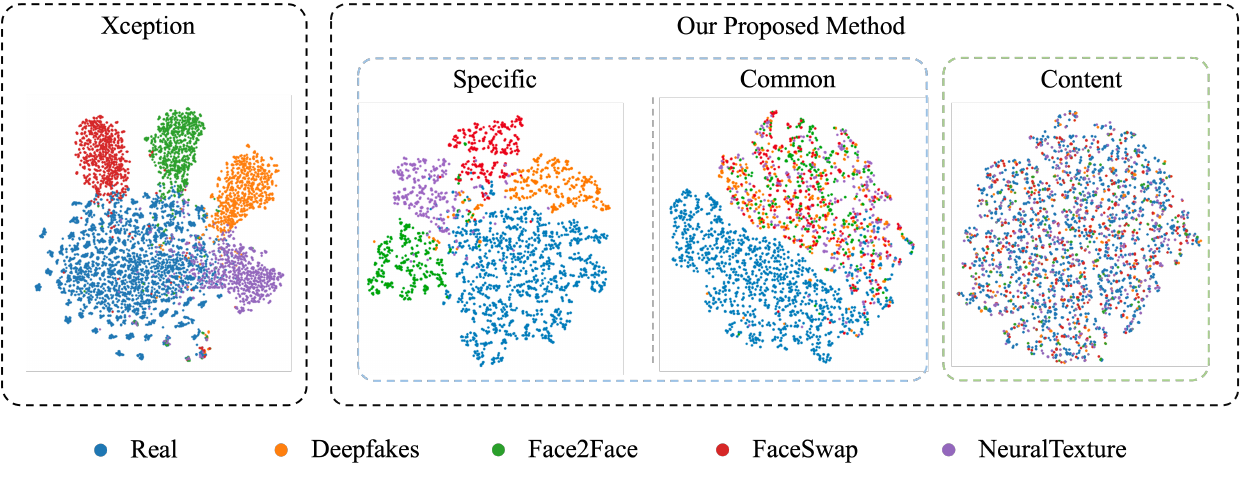} 
      \caption
      {
      The t-SNE~\cite{van2008visualizing} visualization of features extracted from the baseline Xception~\cite{rossler2019faceforensics++} and our framework on FF++~\cite{rossler2019faceforensics++}. In the visualization, images generated by the four methods locate separately in the latent space, which reveals that the baseline Xception actually learns method-specific features, consistent with our forgery-specific module. This observation explains that Xception can mainly recognize specific types of forgeries and thus fail to generalize well to a broader range of forgeries. Additionally, as expected, the common module of our method captures the common forgery features across different methods, while the content module captures only forgery-irrelevant features.
      }
      \label{tsne}
\vspace{-10pt} 
\end{figure*}

Currently, an increasing number of studies are dedicated to tackling the issue of generalization in deepfake detection.
These works typically utilize blending artifacts~\cite{li2020face,shiohara2022detecting} or frequency artifacts~\cite{liu2021spatial,luo2021generalizing}, and some employ adversarial training to synthesize challenging forgeries~\cite{chen2022self}. However, these approaches are limited in their reliance on predefined forgery patterns. For instance, Face X-ray~\cite{li2020face} assumes the presence of a blending region in the forged image, which could potentially curtail the effectiveness of the approach when generalized to novel and unseen forgeries.
In addition, these methods consider the entire feature space when addressing the problem, which could be disrupted by irrelevant factors such as background~\cite{liang2022exploring} and identity~\cite{dong2023implicit}.

To address the above challenges, we adopt the perspective of content and style~\cite{huang2018multimodal} and formulate the problem of deepfake as an integration of two distinct components: content and fingerprint. The content in deepfake refers to elements, \textit{e.g.,} the background, identity, and facial appearance, which are not directly related to the forgery. In contrast, the fingerprint represents the traits that are related to the forgery. The challenge, then, becomes how to effectively disentangle these two components and use only the forgery-related fingerprint for detection.

Several recent studies~\cite{yang2021learning, hu2021improving, liang2022exploring} attempt to address the generalization problem through disentanglement techniques. However, limited generalization capability remains a challenge in many cases. One main reason is the over-reliance on method-specific patterns in most disentanglement methods, which only aim to eliminate the influence of content. However, these methods may still learn patterns that are unique to a specific forgery method, thereby limiting their generalization performance.


To address this issue, we propose a novel disentanglement framework that differs from existing approaches (See Fig.~\ref{concept}). 
Our framework prevents overfitting to both content and specific forgery patterns.
To achieve this, we employ a multi-task disentanglement framework and a conditional decoder to disentangle the input into content and fingerprint components. Moreover, we introduce a contrastive regularization technique to disentangle the fingerprint features into specific and common features. The specific features represent method-specific forgeries, while the common features are shared across different forgery methods. In our approach, only the common features are utilized for detection, which improves the generalization ability of the model. To validate our idea, we conduct a t-SNE visualization~\cite{van2008visualizing} in Fig.~\ref{tsne}, demonstrating that the baseline and our specific components actually learn method-specific texture, while our common components are able to capture the common features across forgeries. Furthermore, the content does not differentiate between real and fake images, as expected.

Our contributions are summarized as follows:

\begin{itemize}
    \item We propose a novel multi-task disentanglement framework to address two main challenges that contribute to the generalization problem in deepfake detection: overfitting to irrelevant features and overfitting to method-specific textures. By uncovering common features, our framework aims to enhance the generalization ability of the model.

    \item We propose a conditional decoder that helps disentangle forgery-irrelevant and forgery features, as well as a contrastive regularization technique that facilitates the disentanglement of common and specific forgery features. By utilizing these technologies, we aim to achieve improved disentanglement.
    
    \item Extensive experiments show that our framework can outperform the performance of current state-of-the-art methods in unseen testing datasets, demonstrating its effectiveness in generalization.
\end{itemize}

\vspace{-8pt}

\section{Related Work}
To date, deepfake detection can be broadly categorized into two types of tasks: image forgery detection~\cite{rossler2019faceforensics++,li2020face,chen2022self} and video forgery detection~\cite{sabir2019recurrent,haliassos2021lips,zheng2021exploring}. This paper specifically focuses on detecting image forgery. 

\vspace{-8pt}

\paragraph{Classical Detection Methods.} 
Conventional deepfake detectors~\cite{afchar2018mesonet, nguyen2019capsule, rossler2019faceforensics++} typically focus on developing optimal CNN architectures. However, these methods often overlook the details present in the frequency domain of fake images, such as compression artifacts. To this end, several works~\cite{qian2020thinking,frank2020leveraging,li2021frequency,gu2022exploiting} utilize frequency information to improve the performance of detectors. Other notable directions are focusing on some specific representations, \textit{i.e.,} forgery region location~\cite{nguyen2019multi}, neuron behaviors~\cite{wang2019fakespotter}, optical flow~\cite{amerini2019deepfake}, landmark geometric features~\cite{sun2021improving}, 3D decomposition~\cite{zhu2021face}, erasing technology~\cite{wang2021representative}, and attentional networks~\cite{dang2020detection,zhao2021multi,wang2022m2tr}. However, the generalization ability towards unseen forgery technologies of these conventional deepfake detectors is still limited.

\vspace{-8pt}

\paragraph{Detection Methods toward Generalization.} Deepfake detection poses a significant challenge in terms of generalization, where detectors perform poorly when training and testing on different data distributions. Despite this challenge, there is a limited amount of research in this area. One early method, FWA~\cite{li2018exposing}, leverages differences in resolution between forgery faces and backgrounds to detect deepfake. Recent works make significant progress in improving the generalization ability. Face X-ray~\cite{li2020face} detects blending boundary artifacts, SPSL~\cite{liu2021spatial} proposes a frequency-based method by phase spectrum analysis, LipForensics~\cite{haliassos2021lips} leverages spatial-temporal networks to identify unnatural mouth movements, SRM~\cite{luo2021generalizing} utilizes the high-frequency noises for generalizable detection, PCL~\cite{zhao2021learning} measures patch-wise similarities of input images to identify deepfake, SBIs~\cite{shiohara2022detecting} and SLADD~\cite{chen2022self} improve generalization ability by combining data augmentation and blending. Although these approaches largely improve the generalization ability of classical detection methods, they are limited by the reliance on predefined forgery patterns and the consideration of the entire feature space, which can be disrupted by unrelated factors such as background~\cite{liang2022exploring} and identity~\cite{dong2023implicit}.

\paragraph{Disentanglement Learning for Deepfake Detection.} 
Disentanglement learning is a method that decomposes complex features into simpler, more narrowly defined variables and encodes them as separate dimensions with high discriminative power~\cite{bengio2013representation,liang2022exploring}.
In the field of deepfake detection, there are relatively few papers that are based on disentanglement learning. These works aim to separate forgery-irrelated and forgery-related features to extract forgery information from variations present in facial images. 
Hu~\emph{et al}.~\cite{hu2021improving} propose a disentanglement framework that separates features, only using the manipulation-related features for detection. Zhang \textit{et al.}~\cite{zhang2020face} go further step by adding additional supervision to improve generalization ability. 
To ensure the independence of the disentangled features, Liang \textit{et al.}~\cite{liang2022exploring} ensure feature independence through content consistency and global representation contrastive constraints.

Despite these efforts to tackle the generalization problem through disentanglement learning, this challenge still exists because these methods only remove the influence of content. In some cases, these methods may still fail to achieve complete disentanglement of forgery features, resulting in overfitting to method-specific textures and thereby limiting their ability to generalize to other unseen forgeries.

\vspace{-4pt}

\section{Methods}

\begin{figure*}[htb]
      \centering 
      \includegraphics[width=0.90\linewidth]{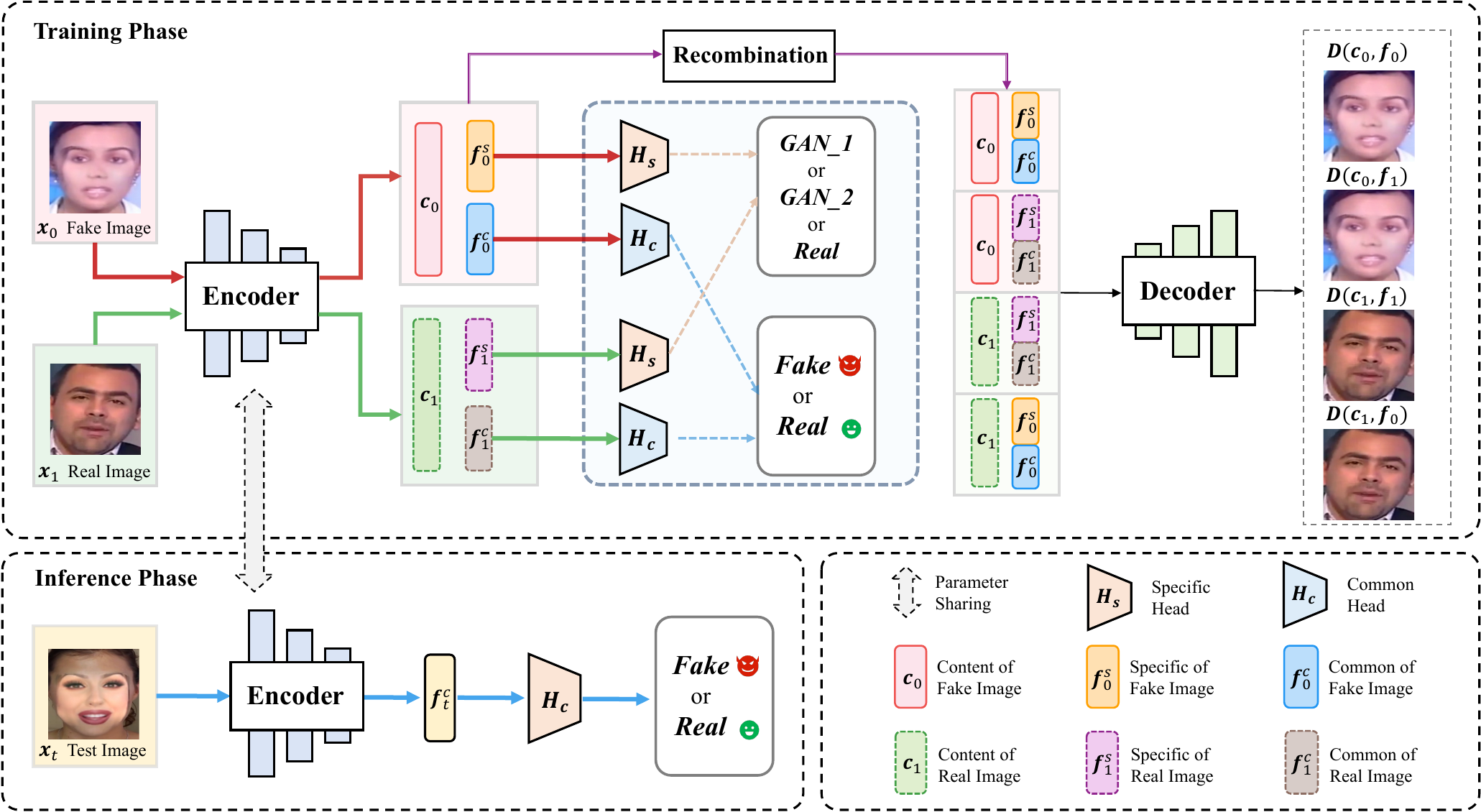} 
      \caption{The overview framework of our proposed method. 1) For the encoder ($\bm{E}$), we utilize it to obtain three distinct components: content, specific fingerprint, and common fingerprint. 2) For the recombination module, we recombine the fingerprints and contents from different input images. 3) For the decoder ($\bm{D}$), we take the fingerprint and content as inputs to generate corresponding reconstruction images. 4) For the classification, we obtain the prediction results of specific and common fingerprints by two different heads ($\bm{H_s}$ and $\bm{H_c}$) to classify the forgery method and determine whether the image is real or fake, respectively.}
      \label{figure:framework}
\end{figure*}

\subsection{Motivation}
There are two main factors that contribute to the generalization problem in deepfake detection. Firstly, many detectors are prone to focus too much on content information that is not directly related to the forgery, \textit{i.e.,} the background, identity, and facial appearance. Secondly, different forgery techniques produce distinct forgery artifacts. These artifacts can be easily detected by a detector that is trained on a specific set of artifacts. However, detectors may be overfitted to one or more specific forgery technologies, leading to a lack of generalization to unseen forgeries. The second problem is often overlooked in previous works.

To address these issues, we propose a multi-task disentanglement learning framework to uncover common features for generalizable deepfake detection. Our framework aims to disentangle the input into the content, specific, and common forgery features. By only utilizing the common forgery features for detection, our framework can help improve the generalization ability of deepfake detectors and avoid the overfitting of both forgery-irrelated and method-specific features. Additionally, we introduce a conditional decoder and a contrastive regularization loss to further aid in disentanglement and enhance the generalization ability of the framework.

\begin{figure}[htb]
      \centering 
      \includegraphics[width=0.9\linewidth]{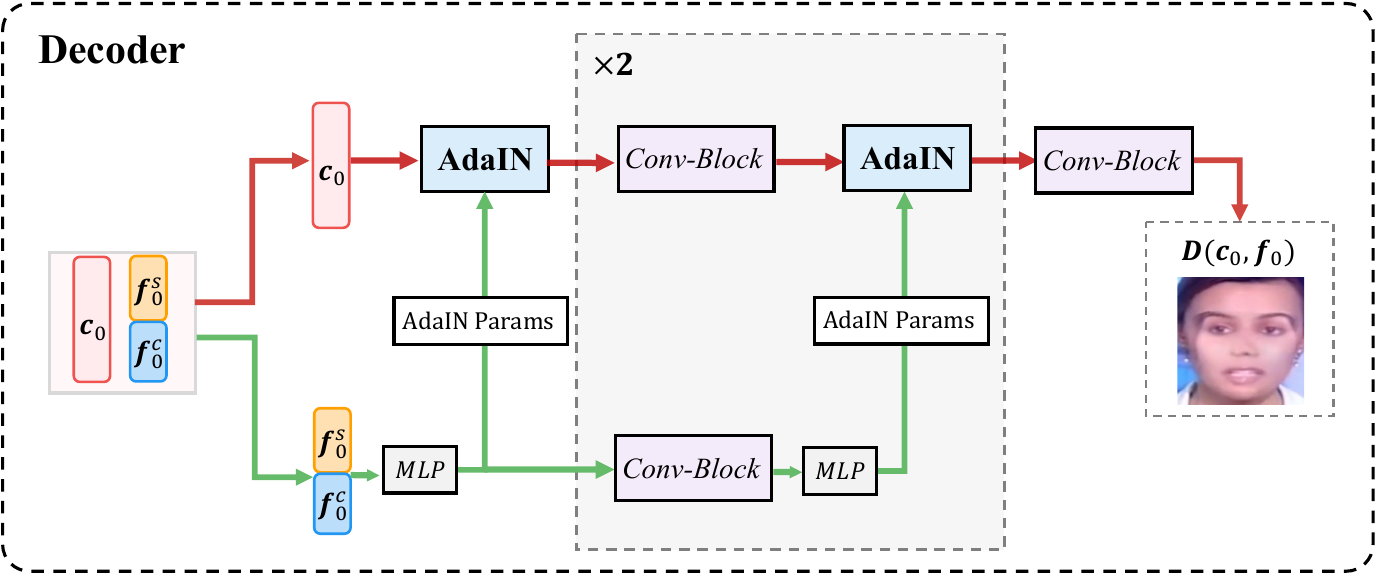} 
      \caption{The architecture of our decoder $\bm{D}$, involves combining the fingerprint and content through AdaIN layers, which are then processed through multiple convolutional blocks along with upsampling layers (indicated as ``Conv-Block" in the figure). The AdaIN layers are utilized twice during this process to fuse the fingerprint as a condition along with the content. Ultimately, the output of the final ``Conv-Block" layer is decoded to reconstruct the image.}
      \label{figure:decoder}
\vspace{-10pt}
\end{figure}

\vspace{8pt}

\subsection{Disentanglement Framework}

Our disentanglement framework, depicted in Fig.~\ref{figure:framework}, consists of an encoder, a decoder, and two classification heads. The encoder comprises a content encoder and a fingerprint encoder that extract content and fingerprint features, respectively. While the two encoders share the same structure, they do not share parameters. The decoder includes multiple convolutional and upsampling layers that reconstruct the image by utilizing fingerprint features as a condition along with content features. The classifier consists of two different heads, one for learning method-specific textures and the other for learning generalizable features across different forgeries. More details about our encoder can be found in the supplementary material.

\subsection{Architecture}

\paragraph{Encoders.}
Our encoder processes a pair of images $(\bm{x}_{0}, \bm{x}_{1})$, where $\bm{x}_{0}$ represents the fake image and $\bm{x}_{1}$ represents the real image. The encoder $\bm{E}$ comprises a content encoder $\bm{E}_c$ and a fingerprint encoder $\bm{E}_f$, extracting the content and forgery features, respectively. We apply the encoder to each pair to obtain the corresponding fingerprint and content features as follows:
\begin{equation} \label{encoder}
\bm{f}_i^{s}, \ \bm{f}_i^{c}, \ \bm{c}_i = \bm{E}(\bm{x}_{i}),
\end{equation}
where $i \in \left\{ 0,1 \right\}$ is the index of the image. Also, $\bm{f}_0^{s}$, $\bm{f}_0^{c}$, $\bm{c}_0$ and $\bm{f}_1^{s}$, $\bm{f}_1^{c}$, $\bm{c}_1$ are denoted by the specific fingerprint, common fingerprint, and content corresponding to each input image pair, respectively.

\paragraph{Decoders.}
Our decoder (See Fig.~\ref{figure:decoder}) reconstructs an image by utilizing its content and fingerprint through a series of upsampling and convolutional layers. Unlike other disentanglement-based deepfake detection frameworks~\cite{zhang2020face,yang2021learning,liang2022exploring} that linearly add forgery and content features for recombination, our decoder applies Adaptive Instance Normalization (AdaIN)~\cite{huang2017arbitrary} for improved reconstruction and decoding, inspired by stylization techniques~\cite{huang2018multimodal}. The AdaIN aligns the mean and variance of the content code to match those of the fingerprint. The formula is written as:
\begin{equation} \label{adain}
\operatorname{AdaIN}(\bm{c}, \bm{f})=\sigma(\bm{f})\left(\frac{\bm{c}-\mu(\bm{c})}{\sigma(\bm{c})}\right)+\mu(\bm{f}),
\end{equation}
where $\bm{c}$ and $\bm{f}$ are the content and the style vectors of the image pair, respectively. The functions $\mu(\cdot)$ and $\sigma(\cdot)$ compute the mean and variance of the input.

\subsection{Objective Function}
To attain disentangled feature representation for detection, we design three distinct loss functions: two classification losses for common and specific forgery features, a contrastive regularization loss for similar and dissimilar image embeddings, and a reconstruction loss to ensure consistency between original and reconstructed images at the pixel level. These losses are combined in a weighted sum to create the overall loss function for training the framework.

\vspace{-5pt}
\paragraph{Multi-Task Classification Loss.}
For classification loss, we propose two different losses of classification. First, we propose a binary classification loss $\mathcal{L}_{ce}^{c}$ computed by the cross-entropy for supervising the model to learn the common feature of different forgery methods:
\begin{equation}
\mathcal{L}_{ce}^{c}=\mathcal{L}_{ce}(\bm{H}_{c}(\bm{f}_{i}^{c}), \ \bm{y}_{i}),
\end{equation}
where $\mathcal{L}_{ce}$ denotes the cross-entropy loss, $\bm{H}_{c}$ is the head for the common forgery feature, which is implemented by several MLP layers. $\bm{y}_{i} \in \left\{\text{fake}, \text{real}  \right\}$ is the binary classification label. In addition, $\mathcal{L}_{ce}^{s}$ is proposed to learn the method-specific patterns by guiding the model to identify which forgery method is applied to the fake image:
\begin{equation}
\mathcal{L}_{ce}^{s}=\mathcal{L}_{ce}(\bm{H}_{s}(\bm{f}_{i}^{s}), \ \bm{y}_{i}^{\prime}),
\end{equation}
where $\bm{H}_{s}$ is the head for specific forgery feature and $\bm{y}_{i}^{\prime} \in \left\{\text{real}, GAN_1, GAN_2, \cdots   \right\}$ donates the label for identifying which instance belong to the image. Note that $\bm{H}_{c}$ and $\bm{H}_{s}$ share the same architecture but not share the parameters.

The multi-task classification loss enables the model to learn both method-specific textures and common features in different forgeries, enhancing the generalization ability of the model.

\paragraph{Contrastive Regularization Loss.}
The objective of contrastive regularization loss is to optimize the similarity and dissimilarity measurements between images.  The contrastive regulation loss is formulated mathematically as:

\begin{equation} \label{con-loss}
\mathcal{L}_{con} = \max \left(\left|\bm{x}_A-\bm{x}_P\right|_2 - \left|\bm{x}_A-\bm{x}_N\right|_2 + \alpha, \ 0 \right),
\end{equation}
where $\alpha$ serves as a margin hyper-parameter.
This method minimizes the Euclidean distance between an anchor image ($\bm{x}_A$) and its similar counterpart ($\bm{x}_P$), while simultaneously maximizing the gap between the anchor image and its dissimilar counterpart ($\bm{x}_N$).
For instance, if $\bm{x}_A$ denotes common features of a genuine image, $\bm{x}_P$ would represent common attributes of another real image, while $\bm{x}_N$ signifies the attributes of a manipulated image.

For the common features, we compute the loss between real and fake images to encourage the model to learn a generalizable representation that across different forgeries. For the specific features, we compute the loss between images of the same forgery to encourage the model to learn method-specific textures for each forgery.

\paragraph{Reconstruction Loss.}
In general, there are two types of reconstruction in our framework: self-reconstruction and cross-reconstruction. For the self-reconstruction, the decoder $\bm{D}$ is applied to the content and fingerprint encoded from the same image to reconstruct the corresponding image, and the formula is written as:

\begin{equation} \label{self-reconstruction}
    \mathcal{L}_{rec}^{s} = \left\|\bm{x}_{0} - \bm{D}(\bm{f}_{0}, \bm{c}_{0})\right\|_1 + \left\|\bm{x}_{1} - \bm{D}(\bm{f}_{1}, \bm{c}_{1})\right\|_1.
\end{equation}
For the cross-reconstruction, we consider the different combinations of fingerprint and content features encoded from the different images. Similarly, the formula of the cross-reconstruction loss is written as:
\begin{equation} \label{cross-reconstruction}
    \mathcal{L}_{rec}^{c} = \left\|\bm{x}_{0} - \bm{D}(\bm{f}_{1}, \bm{c}_{0})\right\|_1 + \left\|\bm{x}_{1} - \bm{D}(\bm{f}_{0}, \bm{c}_{1})\right\|_1.
\end{equation}

Considering both the self-reconstruction and cross-reconstruction loss, the overall image reconstruction loss can be computed as follows:
\begin{equation} \label{L_rec}
	\mathcal{L}_{rec} = \mathcal{L}_{rec}^{s} + \mathcal{L}_{rec}^{c}.
\end{equation}

The image reconstruction loss ensures that the reconstructed image and the original image are consistent at the pixel level. In addition, the two reconstruction losses enhance the disentanglement of features. The self-reconstruction loss penalizes the reconstruction errors by leveraging the latent features of the input image, while the cross-reconstruction loss penalizes the reconstruction errors using the forgery feature and the swapped content features.

\paragraph{Overall Loss.}
The final loss function of the training process is the weighted sum of the above
loss functions.
\begin{equation} \label{overall_loss}
\mathcal{L} = \mathcal{L}_{ce}^{c}+\lambda_{1}\mathcal{L}_{ce}^{s}+\lambda_{2}\mathcal{L}_{rec}+ \lambda_{3}\mathcal{L}_{con},
\end{equation}
where $\lambda_{1},\lambda_{2},\lambda_{3}$ are hyper-parameters for balancing the overall loss.

\begin{table*}
\centering
\scalebox{0.70}{
\begin{tabular}
{c|c|c|c|c|c|c|c|c|c|c|c|c|c|c|c|c}
\toprule
\multirow{2}*{Method}& \multicolumn{3}{c|}{FF-ALL} & \multicolumn{3}{c|}{FF-wo-DF} & \multicolumn{3}{c|}{FF-wo-F2F} & \multicolumn{3}{c|}{FF-wo-FS} & \multicolumn{3}{c|}{FF-wo-NT} & \multirow{2}*{Avg.}\\
\cline{2-16}
& \small{DFDC} & \small{CelebDF} & \small{DFD} & \small{DFDC} & \small{CelebDF} &\small{DFD} & \small{DFDC} & \small{CelebDF} &\small{DFD} & 
\small{DFDC} & \small{CelebDF} &\small{DFD} & \small{DFDC} &\small{CelebDF} &\small{DFD}  \\ 
\midrule
Xception~\cite{rossler2019faceforensics++} &0.651 &0.672 &0.727 &0.651 &0.660 &0.633 &0.646 &0.716 &0.794 &0.665 &0.737 &0.826 &0.647 &0.709 &0.798 &0.702\\
Liang \emph{et al}.~\cite{liang2022exploring} &0.700 &0.706 &0.829 &0.707 &0.699 &0.794 &0.705 &0.698 &0.844 &0.709 &0.713 &0.851 &0.667 &0.672 &0.750 &0.736\\
CORE~\cite{ni2022core}& 0.658 & 0.708 & 0.917 & 0.630 & 0.644 & 0.807 & 0.671 & 0.708 & 0.923 & 0.663 & 0.711 & 0.925 & 0.653 & 0.689 & 0.920 & 0.748 \\
RECCE~\cite{cao2022end} & 0.635 & 0.756 & \underline{0.933} & 0.636 & 0.604 & 0.821 & 0.661 & 0.724 & \textbf{0.930} & 0.651 & 0.778 & \underline{0.928} & 0.659 & 0.754 & 0.932 & 0.760 \\
FWA~\cite{li2018exposing} & 0.650 & 0.755 & 0.870 & 0.635 & \textbf{0.771} & \underline{0.885} & 0.701 & 0.778 & 0.924 & 0.689 & 0.752 & 0.850 & 0.670 & 0.744 & 0.875 & 0.770 \\
Face X-ray~\cite{li2020face} &0.710 &0.740 &0.890 &0.726 &0.668 &0.838 &0.734 &0.716 &0.899 &0.705 &0.693 &0.907 &0.731 &0.731 &0.901 &0.773\\
SLADD~\cite{chen2022self} & 0.751 & 0.753 & 0.900 & 0.738 & 0.705 & 0.841 & 0.757 & 0.754 & 0.815 & \underline{0.713} & 0.733 & 0.883 & 0.754 & 0.741 & 0.894 & 0.782 \\
F3Net~\cite{qian2020thinking} &0.743 &0.668 &0.926 &0.748 &0.648 &0.872 &\textbf{0.765} &0.692 &0.914 &\textbf{0.719} &0.680 &0.925 &0.779 &0.760 &\underline{0.933} &0.785\\
SRM~\cite{luo2021generalizing} &\underline{0.771} &0.770 &0.915 &0.743 &0.746 &0.874 &0.745 &0.757 &0.909 &0.699 &0.768 &0.923 &0.778 &0.779 &0.891 &0.805\\
SPSL~\cite{liu2021spatial} & 0.742 & \underline{0.787} & 0.927 & \underline{0.749} & \underline{0.753} & \textbf{0.898} & \underline{0.759} & \textbf{0.797} & \underline{0.929} & 0.664 & \underline{0.794} & 0.924  & \underline{0.797} & \textbf{0.813} & 0.924 & \underline{0.817} \\

\midrule

Ours &\textbf{0.805} &\textbf{0.824} &\textbf{0.945} &\textbf{0.767} &0.749 &0.870 &\textbf{0.765} &\underline{0.782} &0.908 &0.711 &\textbf{0.800} &\textbf{0.943} &\textbf{0.800} &\underline{0.808} &\textbf{0.943} &\textbf{0.828}\\
\bottomrule
\end{tabular}}
\vspace{0.1 cm}
\caption{Comparisons of generalization ability with competing methods implemented by ourselves. We use two different data configurations: ``FF-ALL", which includes all data generated by four forgeries, and ``FF-wo-DF", ``FF-wo-F2F", ``FF-wo-FS", and ``FF-wo-NT", which use the FF++ dataset but drop DF, F2F, FS, and NT, respectively. The best results are highlighted in bold font, while the second-best results are underlined.}
\label{tab result}
\vspace{-0.3 cm}
\end{table*}

\section{Experiments}

\subsection{Settings}
\paragraph{Datasets.}
To evaluate the generalization ability of the proposed framework, our experiments are conducted on four large-scale benchmark databases: FaceForensics++ (FF++)~\cite{rossler2019faceforensics++}, DeepfakeDetection (DFD)~\cite{dfd}, Deepfake Detection Challenge (DFDC)~\cite{dfdc}, and CelebDF~\cite{li2019celeb}. FF++~\cite{rossler2019faceforensics++} is a large-scale database comprising more than 1.8 million forged images from 1000 pristine videos. Forged images are generated by four face manipulation algorithms using the same set of pristine videos, \textit{i.e.,} DeepFakes (DF)~\cite{deepfake}, Face2Face (F2F)~\cite{thies2016face2face}, FaceSwap (FS)~\cite{faceswap}, and NeuralTexture (NT)~\cite{thies2019deferred}. 
To evaluate the generalization ability of our framework, we follow prior research works~\cite{li2020face,chen2022self} and conduct experiments on three widely used face-manipulated datasets, \textit{i.e.,} DFDC~\cite{dfdc}, CelebDF~\cite{li2019celeb}, and DFD~\cite{dfd}. 
Note that there are three versions of FF++ in terms of compression level, \textit{i.e.,} raw, lightly compressed (HQ), and heavily compressed (LQ). Since realistic forgeries often have a limited quality, the HQ and LQ versions are used in experiments. Following previous works~\cite{li2020face,chen2022self}, \textbf{the HQ version of FF++ is adopted by default.} If any deviation from this default, it will be explicitly stated.

\vspace{-8pt}
\paragraph{Implementation.}
We use a modified version of Xception~\cite{rossler2019faceforensics++} as the backbone network, with model parameters initialized by pre-training on ImageNet. Face extraction and alignment are performed using DLIB~\cite{sagonas2016300}. Following previous works~\cite{chen2022self}, the aligned faces are resized to $256\times 256$ for both the training and testing. We use the Adam~\cite{kingma2014adam} for optimization with the learning rate of 0.0002, and the batch size is fixed as 32.
In the overall loss function in Eq.~(\ref{overall_loss}), we set $\lambda_{1}$ to $\lambda_{3}$ as 0.1, 0.3, 0.05 empirically. The margin $\alpha$ in Eq.~(\ref{con-loss}) is set to 3. We also apply some widely used data augmentations, \textit{i.e.,} image compression, horizontal flip, and random brightness contrast.

\vspace{-8pt}
\paragraph{Evaluation Metrics.} 
We report the Area Under Curve (AUC) metric to compare our proposed method with prior works, which is consistent with the evaluation approach adopted in many previous works~\cite{li2020face,qian2020thinking,rossler2019faceforensics++,luo2021generalizing,chen2022self}.
\textbf{The default evaluation metric employed is the AUC.}
We also report other metrics such as Accuracy (ACC), Average Precision (AP), and Equal Error Rate (EER) for a more comprehensive evaluation of our method. Please refer to our supplementary for more details.


\subsection{Generalization Ability Evaluation}
\paragraph{Comparison with competing methods.}
To assess the generalization capacity of our framework, \textbf{we reproduce ten competing methods under consistent conditions for a comprehensive comparison:}
Xception~\cite{rossler2019faceforensics++}, Face X-ray~\cite{li2020face}, F3Net~\cite{qian2020thinking}, SRM~\cite{luo2021generalizing}, SPSL~\cite{liu2021spatial}, RECCE~\cite{cao2022end}, CORE~\cite{ni2022core}
, SLADD~\cite{chen2022self}, and Liang \textit{et al.}~\cite{liang2022exploring}. 
We use the provided codes of Xception, RECCE, SLADD, CORE, FWA, and SRM from the authors. We reimplement Face X-ray, F3Net, SPSL, and Liang \textit{et al.}~\cite{liang2022exploring} rigorously following the companion paper’s instructions and train these models under the same settings. 

We conduct this experiment by training the models on the FF++~\cite{rossler2019faceforensics++} and then evaluate these models in DFD~\cite{dfd}, DFDC~\cite{dfdc}, and CelebDF~\cite{li2019celeb}, respectively. This setting is challenging in generalization ability evaluation since the testing sets are collected from different sources and share much less similarity with the training set. 

The results of the comparison between different methods are presented in Tab.~\ref{tab result}, which shows the performance in terms of the AUC metric. It is evident that the proposed disentanglement framework and multi-task learning strategy lead to superior performance compared to other models in most cases, achieving the overall best results. 

Liang \textit{et al.}~\cite{liang2022exploring} proposes a disentanglement framework for content information removal, but their model is still prone to overfitting to method-specific patterns, leading to the limitation of the generalization. On the contrary, the disentanglement framework we proposed is designed to learn generalizable features across different forgeries by the multi-task learning strategy, thereby achieving improved generalization performance. 

Face X-ray~\cite{li2020face} and FWA~\cite{li2018exposing} use blended artifacts in forgeries to achieve generalization. However, these two methods have limited generalization ability when the patterns in the training and testing datasets differ. This is because Face X-ray learns to identify the boundary patterns that are sensitive to the post-processing operations varying in different datasets. On the contrary, our proposed framework learns common representations that are not dependent on specific post-processing operations. 

SRM~\cite{luo2021generalizing}, SPSL~\cite{liu2021spatial}, and F3Net~\cite{qian2020thinking} utilize frequency components of images to distinguish between forgeries and pristine images. However, the experimental results show that their generalization performance is inferior to the proposed approach. This could be due to the fact that these frequency cues that are effective on the FF++ may not generalize to other datasets with different post-processing steps.

CORE~\cite{ni2022core}, RECCE~\cite{cao2022end}, SLADD~\cite{chen2022self} are recent detectors that focus on different detection algorithms: loss design, reconstruction learning, and adversarial training. However, these detectors could still be disrupted by unrelated factors such as race, gender, or identity because they operate in the entire feature space, which inevitably includes these unrelated aspects (similarly indicated in previous work~\cite{zhang2020face}). From Tab.~\ref{tab result}, our UCF (82.8\%) largely outperforms SLADD (78.2\%) in terms of the average AUC.

Finally, Xception~\cite{rossler2019faceforensics++} serves as a CNN baseline and does not incorporate any augmentation, disentanglement, feature engineering, or frequency information. Its performance drops dramatically in the case of unseen forgeries, highlighting the importance of incorporating these techniques in face forgery detection models.




    \begin{table}[tb!]
      \centering
      \caption{Comparison with state-of-the-art methods on CelebDF and DFDC. The results of other works are mainly cited from \cite{chen2022self,yu2022improving,wang2023dynamic}}
      \vspace{-5pt}
      \scalebox{0.83}{
      \begin{tabular}{c|c|c|c} \toprule
        Model& Training Set & CelebDF & DFDC\\ 
        \midrule
        Two-stream~\cite{zhou2017two} &FF++&0.538 &-\\
        Meso4~\cite{afchar2018mesonet} &Self-made&0.548&0.497\\
        MesoInception4~\cite{afchar2018mesonet} &Self-made&0.536&0.499\\
        DSP-FWA~\cite{li2018exposing} &FF++&0.646&0.646\\
        VA-MLP~\cite{matern2019exploiting} &FF++&0.550 &-\\
        Multi-task~\cite{nguyen2019multi} &FF++&0.543 &-\\
        Headpose~\cite{yang219exposing} &UADFV&0.546 & -\\
        Capsule~\cite{nguyen2019capsule} &FF++&0.575 &0.575\\
        SMIL~\cite{li2020sharp} &FF++&0.563 &0.563\\
        Two-branch~\cite{masi2020two} &FF++&0.734 &0.734\\
        Schwarcz \emph{et al}.~\cite{schwarcz2021finding} &FF++&0.667 &0.673 \\
        PEL~\cite{gu2022exploiting} &FF++&0.692 &0.633\\
        MADD~\cite{zhao2021multi} &FF++&0.674  &-\\
        Local-relaion~\cite{chen2021local} &FF++&0.783  &0.765\\
        CFFs~\cite{yu2022improving} &FF++ &0.742 &0.721 \\
        Zhuang \emph{et al}.~\cite{zhuang2022towards} &FF++&0.728 & - \\
        SFDG~\cite{wang2023dynamic} &FF++&0.758 & 0.736 \\
        \hline
        Ours & FF++ & \textbf{0.824} & \textbf{0.805}\\
        
        \bottomrule
      \end{tabular}
      }
      \label{tab:cmp_sota}
    \end{table}

\begin{table}[tb!]
      \centering
      \caption{Comparison on FF++ with methods using disentanglement learning. }
      \scalebox{0.75}{
      \begin{tabular}{c|c|c|c|c|c}
        \toprule
        \multirow{2}*{Training} & \multirow{2}*{Method}
        &\multicolumn{4}{c}{Testing AUC}\\
        \cmidrule(lr){3-6}
        ~&~&
        FF++(LQ) & CelebDF & DFD & DFDC \\
        \midrule
        
        \multirow{3}*{FF++}
        & Xception~\cite{rossler2019faceforensics++}
        & 0.683 & 0.672 & 0.727 & ~ 0.651\\
        
        ~&Liang \emph{et al}.~\cite{liang2022exploring} & 0.714 & 0.706 & 0.829& ~ 0.700\\

        ~&Ours & \textbf{0.833}&  \textbf{0.824}&  \textbf{0.945}& ~ \textbf{0.805}\\
         \bottomrule
      \end{tabular}}
      \label{tab:disentanglement}
      \vspace{-5pt} 
    \end{table}

\vspace{-8pt}

\paragraph{Comparison with disentanglement-based methods.}
For disentanglement-based detection frameworks, we identify one prior work, Liang \textit{et al.}~\cite{liang2022exploring}, that shares similarities with our approach. Their framework aims to remove content information and introduces two modules to enhance the independence of disentangled features.
To ensure a fair comparison, we carefully implement their framework by following the settings of the original paper, as it is not available as an open-source resource. We train the baseline Xception, Liang \textit{et al.}~\cite{liang2022exploring}, and ours on the FF++ and evaluated them on FF++ (LQ), CelebDF, DFD, and DFDC. As reported in Tab.~\ref{tab:disentanglement}, we observe that Liang \textit{et al.}~\cite{liang2022exploring} improve upon the baseline largely, demonstrating the essential of removing content information. Additionally, UCF outperforms Liang \textit{et al.}~\cite{liang2022exploring} on all testing datasets, showing the efficacy of uncovering common features.
%

\vspace{-8pt}

\paragraph{Comparison with state-of-the-art methods.} 
We further evaluate our method against other state-of-the-art models.
%
The results, as shown in Tab.~\ref{tab:cmp_sota}, demonstrate the effective generalization ability of our framework as it outperforms other methods, achieving the best performance in terms of the AUC metric on both CelebDF and DFDC. The results of some methods are directly cited
from ~\cite{chen2022self,yu2022improving, wang2023dynamic}.
\textbf{Following a comprehensive evaluation against 27 state-of-the-art detectors (10 implemented in this study and 17 referenced), we demonstrate the robust generalization capability of our proposed framework.}
It is worth noting that both UCF and Zhuang~\emph{et al}.~\cite{zhuang2022towards} aim to tackle the challenging issue of overfitting to method-specific artifacts. However, their technical methodologies are totally different (disentangle vs. adversarial learning). Actually, we offer a fresh and distinct solution to this problem. Moreover, UCF (82.4\%) significantly outperforms Zhuang~\emph{et al}.~\cite{zhuang2022towards} (72.8\%) on CelebDF in terms of AUC.


\subsection{Ablation Study}
\begin{table}[tb!]
      \centering
      \caption{
      Ablation study regarding the effectiveness of our disentanglement framework, multi-task learning strategy, and contrastive regularization loss. ``D'' and ``M'' represent our basic disentanglement framework and the multi-task learning module, respectively. ``C'' represents the contrastive learning module. Results in gray indicate the within-dataset performance.
    }
      \scalebox{0.75}{
      \begin{tabular}{c|c|c|c|c|c}
        \toprule
        \multirow{2}*{Training} & \multirow{2}*{Method}
        &\multicolumn{4}{c}{Testing AUC}\\
        \cmidrule(lr){3-6}
        ~&~&
        FF++ & CelebDF & DFD & DFDC \\
        \midrule
        
        \multirow{4}{*}{\centering FF++}
        & Xception~\cite{rossler2019faceforensics++}
        & \cellcolor{Gray}{0.986} & 0.672 & 0.727 & ~ 0.651\\
        
        ~&Xception + D & \cellcolor{Gray}{0.995} & 0.785 & 0.933& ~ 0.772\\
        
        ~&Xception + D + M & \cellcolor{Gray}{0.995}&  0.804&  0.944& ~ 0.785\\

        ~&Xception + D + M + C & \cellcolor{Gray}{\textbf{0.996}}&  \textbf{0.824}&  \textbf{0.945}& ~ \textbf{0.805}\\
         \bottomrule
      \end{tabular}}
      \label{tab:ablation}
      \vspace{-5pt} 
    \end{table}

\vspace{5mm}

        
        


        

\vspace{-10pt}

\paragraph{Effects of our disentanglement framework and multi-task learning strategy.}
To evaluate the impact of the proposed disentanglement framework and multi-task learning strategy on generalization ability, we conduct an ablation study on several datasets. Specifically, we train all models on FF++ and evaluate their performance on FF++~\cite{rossler2019faceforensics++}, DFD~\cite{dfd}, DFDC~\cite{dfdc}, and CelebDF~\cite{li2019celeb}. The results are reported in Tab.~\ref{tab:ablation} using the AUC metric.
The evaluated variants include the baseline Xception, Xception with the proposed disentanglement framework (Xception + D), the proposed disentanglement framework with the multi-task learning strategy (Xception + D + M), and the multi-task disentanglement framework with the contrastive regularization (Xception + D + M + C).

Regarding the ablation study, we observed the following.
Firstly, the four variants achieve relatively similar results on FF++ (within-dataset evaluation).
Secondly, implementing the basic disentanglement framework leads to a significant improvement in DFD, DFDC, and CelebDF (cross-dataset evaluation), indicating the generalization ability is improved largely when applying the proposed disentanglement framework for the content information removal. 
Thirdly, the multi-task disentanglement outperforms the basic disentanglement, indicating that the multi-task learning strategy is effective in improving the generalization ability of the model.
Finally, combining the proposed multi-task disentanglement with the contrastive regularization loss achieves the best results in both within-dataset and cross-dataset evaluations, supporting the effectiveness of each module.

\begin{table}[tb!]
      \centering
      \caption{
      Ablation study regarding the effectiveness of the conditional decoder. “CD” represents that we use the conditional decoder for image reconstruction. Otherwise, we use the linearly add for combining the fingerprint and content. Results in gray indicate the within-dataset performance.
    }
      \scalebox{0.8}{
      \begin{tabular}{c|c|c|c|c|c}
        \toprule
        \multirow{2}*{Training} & \multirow{2}*{Method}
        &\multicolumn{4}{c}{Testing AUC}\\
        \cmidrule(lr){3-6}
        ~&~&
        FF++ & CelebDF & DFD & DFDC \\
        \midrule
        
        \multirow{2}{*}{\centering FF++}
        & Ours
        & \cellcolor{Gray}{0.995} & 0.811 & 0.932 & ~ 0.786\\

        ~& Ours + CD & \cellcolor{Gray}{\textbf{0.996}}&  \textbf{0.824}&  \textbf{0.945}& ~ \textbf{0.805} \\
         \bottomrule
      \end{tabular}}
      \label{tab:decoder}
      \vspace{-5pt} 
    \end{table}

\begin{table}[tb!]
      \centering
      \caption{
      Comparison with binary classification results of different forgery features.
    }
      \scalebox{0.8}{
      \begin{tabular}{c|c|c|c|c|c}
        \toprule
        \multirow{2}*{Training} & \multirow{2}*{Method}
        &\multicolumn{4}{c}{Testing AUC}\\
        \cmidrule(lr){3-6}
        ~&~&
        FF++ & CelebDF & DFD & DFDC \\
        \midrule
        
        \multirow{4}{*}{\centering FF++}
        & Xception~\cite{rossler2019faceforensics++}
        & \cellcolor{Gray}{0.986} & 0.672 & 0.727 & ~ 0.651\\

        ~& Specific Forgery & \cellcolor{Gray}{0.987}&  0.681&  0.842& ~ 0.667 \\
        
        ~& Whole Forgery & \cellcolor{Gray}{0.995} & 0.785 & 0.933& ~ 0.772\\

        ~& Common Forgery & \cellcolor{Gray}{\textbf{0.996}}&  \textbf{0.824}&  \textbf{0.945}& ~ \textbf{0.805} \\
         \bottomrule
      \end{tabular}}
      \label{tab:binary}
      \vspace{-10pt} 
    \end{table}

\vspace{-8pt}

\paragraph{Effects of our conditional decoder.}
In contrast to other disentanglement-based detection frameworks~\cite{zhang2020face,yang2021learning,liang2022exploring} that use linear addition to combine fingerprint and content features for recombination, our proposed decoder utilizes AdaIN~\cite{huang2017arbitrary} to incorporate the fingerprint as a condition with the content for improved reconstruction and decoding. To evaluate the impact of the conditional decoder on the generalization ability, we conduct an ablation study on the proposed framework with and without the conditional decoder.  Results in Tab.~\ref{tab:decoder} demonstrate that our proposed conditional decoder can achieve improved performance on both within- and cross-datasets, highlighting the importance of using AdaIN layers for reconstruction and decoding.

\vspace{-8pt}

\paragraph{Comparison with binary classification results of different forgery features.}
To evaluate the effectiveness of the proposed multi-task learning strategy, binary classification results are compared based on the common, specific, and whole forgery features. Tab.~\ref{tab:binary} shows that the common features exhibit superior generalization performance compared to the specific features. The comparison of the common and whole forgery features reveals that the whole forgery features are not as effective as the common features, mainly due to the presence of specific features, which may lead to overfitting to method-specific textures. 




\begin{table}[tb!]
      \centering
      \caption{
      Comparing the performance of the baseline and our proposed framework using different backbones. The best result is highlighted in bold font. ``Avg." represents the average AUC for cross-datasets.
    }
      \scalebox{0.69}{
      \begin{tabular}{c|c|c|c|c|c|c}
        \toprule
        \multirow{2}*{Training} & \multirow{2}*{Method}
        &\multicolumn{5}{c}{Testing AUC}\\
        \cmidrule(lr){3-7}
        ~&~&
        FF++(LQ) & CelebDF & DFD & DFDC & Avg. \\
        \midrule
        
        \multirow{4}{*}{\centering FF++}
        
    & Xception~\cite{rossler2019faceforensics++}
    & 0.683 & 0.672 & 0.727 & 0.651 & 0.683\\
    
    ~& Ours (Xception)
    & 0.833 & 0.824 & 0.945 & \textbf{0.805} & 0.852\\

    ~& ConvNext~\cite{liu2022convnet}
    & 0.779 & 0.788 & 0.912 & 0.753  & 0.808\\
    
    ~& Ours (ConvNext)
    & \textbf{0.845} & \textbf{0.869} & \textbf{0.946} & 0.802 & \textbf{0.866}\\

         \bottomrule
      \end{tabular}}
      \label{tab:convnext}
      \vspace{-10pt} 
    \end{table}

\vspace{5mm}



\vspace{-15pt}

\begin{figure}[htb]
      \centering 
      \includegraphics[width=0.9\linewidth]{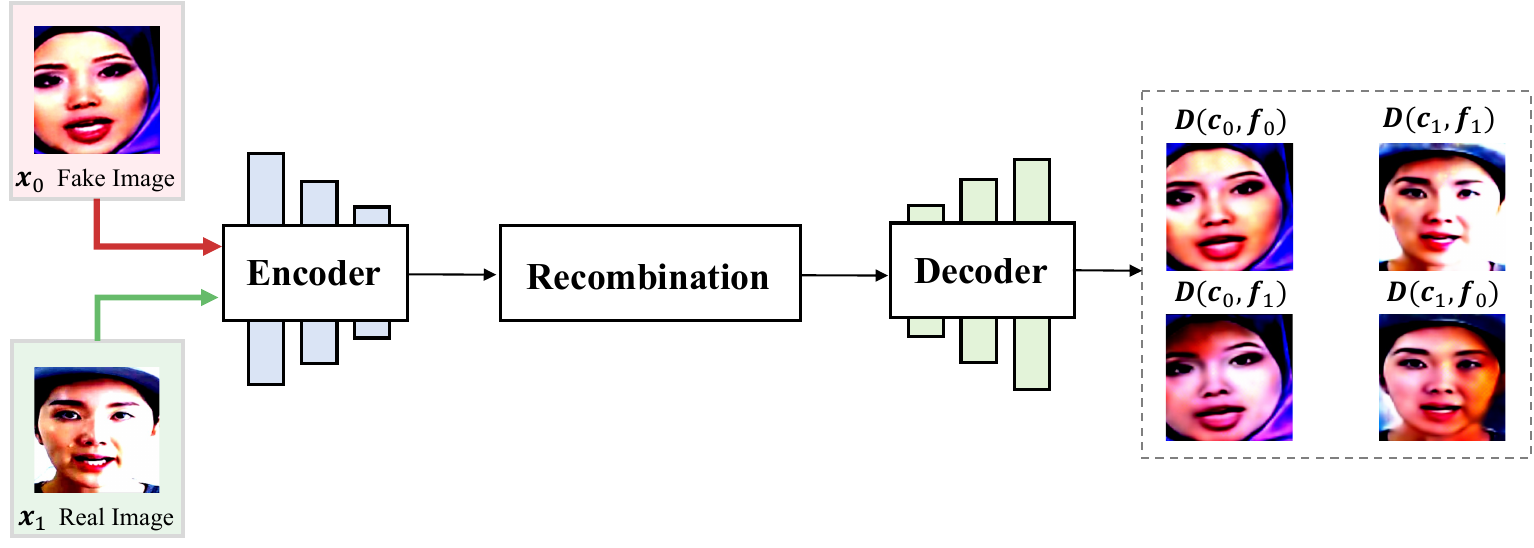} 
      \caption{Visualization of the reconstruction images during the training process.}
      \label{fig:reconstruction}
\vspace{-10pt}
\end{figure}

\paragraph{Exploring Generalization performance of different backbone choices.}
In this section, we investigate the choice of backbone on the generalization ability of our proposed framework. We use Xception as the backbone in our previous experiments to align with other related works, but our multi-task framework is not limited to this choice. To evaluate the effectiveness of our framework with different backbones, we adopt a recent SOTA backbone ConvNeXt~\cite{liu2022convnet} and conduct an ablation study. The results of the ablation study, shown in Tab.~\ref{tab:convnext}, demonstrate that our proposed framework can largely improve the generalization performance of both Xception and ConvNeXt backbones. This suggests that our framework is effective and applicable to different backbone choices.
Additionally, to further highlight the plug-and-play nature of our proposed framework, we extend its application to ResNet~\cite{he2016deep} and EfficientNet~\cite{tan2019efficientnet} backbones. More details can be accessed in our supplementary materials.

\subsection{Visualization}

\paragraph{Visual examples of reconstructed images.}
Within the framework we propose, the generation of reconstruction images occurs during the recombination phase of training. These images serve a pivotal role in ensuring the effective disentanglement of content and fingerprint features, as depicted in Fig.~\ref{fig:reconstruction}.

The content features within our framework are specifically designed to capture appearance, identity, gender, and other forgery-related features. In the visual examples of the reconstructed images (see Fig.~\ref{fig:reconstruction}), it is evident that the images sharing the same content code exhibit a marked similarity. This resemblance persists even when the fingerprint features, which represent unique identifiers separate from the content, are derived from other individuals. The observed similarity in the reconstructed images with identical content codes substantiates the efficacy of our framework in accurately isolating content features from other elements of the image.
Contrary to the content features, the fingerprint features do not alter the content information of the original inputs. However, a close examination of the reconstructed images with the same content code but different fingerprint codes reveals subtle differences.





\section{Conclusion}
In this paper, we propose a novel disentanglement framework that can generalize well in unseen deepfake datasets. Our approach is grounded in the idea that a generalizable deepfake detector should be able to capture the generalizable features across different types of forgeries. To this end, we introduce a multi-task disentanglement framework to uncover the common features. Additionally, we also introduce a conditional decoder and a contrastive regularization loss to enhance the disentanglement process. In this manner, the model can avoid overfitting to forgery-irrelevant and method-specific forgery textures, leading to a more generalizable detector. 
To evaluate the effectiveness of our proposed method, we conduct extensive experiments on several benchmark datasets and compare our results against existing state-of-the-art methods. Overall, our proposed framework represents a promising step toward the development of more generalizable deepfake detectors.


\noindent\textbf{Ethics Statement.} All facial images utilized in this work are sourced from publicly available datasets and are appropriately attributed through proper citations. Our research adheres to strict ethical guidelines throughout the experimental process. There is no compromise on personal privacy during the experiments conducted in this work.

\paragraph{Acknowledgment.}
Baoyuan Wu was supported by the National Natural Science Foundation of China under grant No. 62076213, Shenzhen Science and Technology Program under the grants: No. RCYX20210609103057050, No. ZDSYS20211021111415025, No. GXWD20201231105722002-20200901175001001.

\clearpage

{\small
\bibliographystyle{ieee_fullname}
\bibliography{egbib}
}

\end{document}